\pdfoutput=1
\documentclass[11pt]{article}

\usepackage[]{ACL2023}

\usepackage{times}
\usepackage{latexsym}

\usepackage{algorithm}
\usepackage[noend]{algorithmic}
\usepackage[center]{subfigure}
\usepackage{color,graphicx}
\usepackage{mathtools}
\usepackage{amsmath}
\usepackage{amssymb}
\usepackage{url}
\usepackage{multicol}
\usepackage{multirow}
\usepackage{booktabs}
\usepackage{pifont}
\newcommand{\cmark}{\ding{51}}%
\newcommand{\xmark}{\ding{55}}%

\usepackage[T1]{fontenc}
\usepackage[utf8]{inputenc}

\usepackage{microtype}

\usepackage{inconsolata}

\title{ActPlan-1K: Benchmarking the Procedural Planning Ability of Visual Language Models in Household Activities}

\author{First Author \\
  Affiliation / Address line 1 \\
  Affiliation / Address line 2 \\
  Affiliation / Address line 3 \\
  \texttt{email@domain} \\\And
  Second Author \\
  Affiliation / Address line 1 \\
  Affiliation / Address line 2 \\
  Affiliation / Address line 3 \\
  \texttt{email@domain} \\}

\author{{\bf Ying Su}$^1$, {\bf Zhan Ling}$^2$, {\bf Haochen Shi}$^1$, {\bf Jiayang Cheng}$^{1}$, \\ {\bf Yauwai Yim}$^1$,
{\bf Yangqiu Song}$^1$ \\
$^{1}$HKUST, $^{2}$University of California, San Diego \\
 \texttt{ysuay@connect.ust.hk}, \texttt{z6ling@ucsd.edu} \\
 \texttt{hshiah@connect.ust.hk, \{jchengaj,ywyimaa,yqsong\}@cse.ust.hk} \\
\\
}

\begin{document}
\maketitle

\begin{abstract}
Large language models~(LLMs) have been adopted to process textual task description and accomplish procedural planning in embodied AI tasks because of their powerful reasoning ability. 
However, there is still lack of study on how vision language models~(VLMs) behave when multi-modal task inputs are considered.
Counterfactual planning that evaluates the model's reasoning ability over alternative task situations are also under exploited. 
In order to evaluate the planning ability of both multi-modal and counterfactual aspects, we propose ActPlan-1K.
ActPlan-1K is a multi-modal planning benchmark constructed based on ChatGPT and household activity simulator iGibson2.
The benchmark consists of 153 activities and 1,187 instances. 
Each instance describing one activity has a natural language task description and multiple environment images from the simulator. The gold plan of each instance is action sequences over the objects in provided scenes. 
Both the correctness and commonsense satisfaction are evaluated on typical VLMs.
It turns out that current VLMs are still struggling at generating human-level procedural plans for both normal activities and counterfactual activities.
We further provide automatic evaluation metrics by finetuning over BLEURT model to facilitate future research on our benchmark.
\end{abstract}

%To further measure if the VLMs can handle comlicated situated circumstances in real-life scenarios, we also design counterfactual activities based on the normal activities.
\section{Introduction}
Recent researches adopt large language model (LLM) or multi-modal large language model (MLLM) as agents to accomplish embodied AI tasks \cite{ahn2022can, wang2023voyager, shah2022lmnav, li2022pre, driess2023palm, huang23vlmaps, zhang2023vln, xi2023rise}. The LLMs generate procedural plan according to textual descriptions or embodied observations. The procedural plan consists of action sequences that needs to achieve in order to finish the task. 

While LLMs can generate plausible action plans over current embodied AI tasks with their powerful reasoning ability on texts, it is unclear of how VLM agents behave on multi-modal embodied AI tasks. Previous works typically evaluate
activity completeness in simulator environments \cite{2019ALFRED, puig2020watch, 2021BEHAVIOR}, lacking of evaluation on procedural plans over multi-modal task descriptions. 

%Moreover, previous household activity benchmarks lacks evaluation of counterfactual circumstances which requires counterfactual reasoning ability over various alternative situations \cite{Stalnaker1996knowledge}, are also unexploited.  lacking of special design on benchmarking counterfactual activities.

%Among the entire pipeline, LLM or VLMs mainly take in the natural language task instruction and generate admissible low-level action plans, serving as the role of procedural planner \cite{ahn2022can, wang2023voyager}. The quality of the plan greatly influence the completeness and final success of the task. Previous works either evaluates the completeness of final goal \cite{2021BEHAVIOR, pmlr-v205-li23a} or simple activities \cite{puig2018virtualhome, 2019ALFRED}, unable to evaluate the planning ability of VLMs systematically on complicated activities of human level.

\begin{figure}[t]
\centering
\includegraphics[scale=0.55, trim={0cm 0 0cm 0.5cm}]{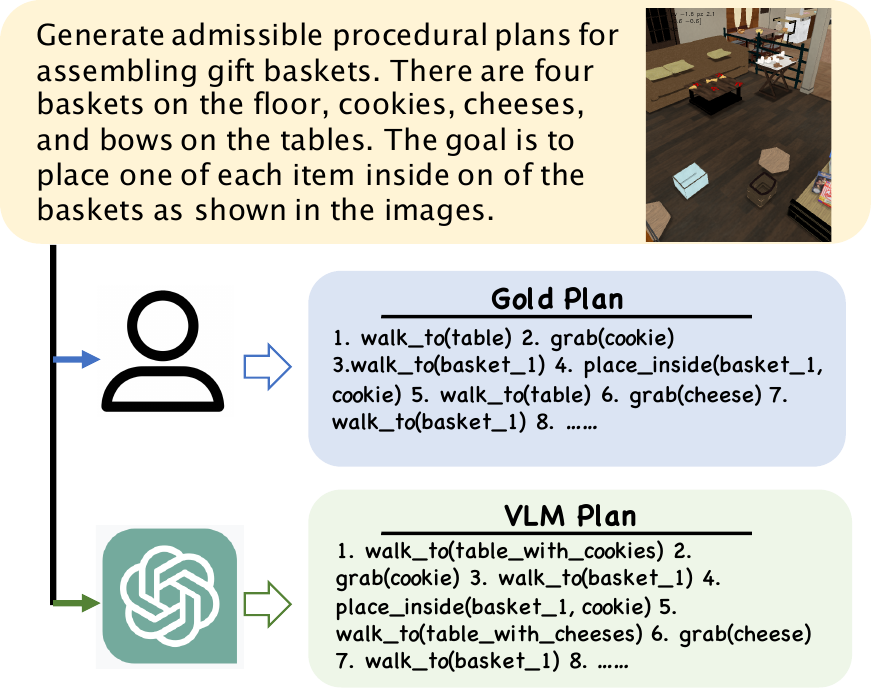}
\caption{Generating procedural plan for household activities with VLMs via prompting with task description and environment images. }
\vskip-1.0em
\label{fig:intro}
\end{figure}

In this paper, we address the rare evaluation on the procedural planning ability of VLMs for human activities by proposing a new benchmark ActPlan-1K. The multi-modal benchmark consists of human activity descriptions in natural language and environment images, mimicking the real-world household scenarios. An example is presented in Figure \ref{fig:intro}. By taking in the multi-modal inputs, VLMs generate admissible procedural plan composed by actions and objects.

\begin{table*}[t]
\resizebox{\textwidth}{!}{
    \begin{tabular}{p{5cm}p{2.8cm}p{1.7cm}p{1.5cm}p{2cm}p{2cm}p{2cm}p{2cm}p{2cm}}
    \toprule
     & ActPlan-1K  & EgoPlan-Bench & COPLAN & Behavior-100 & AI2Thor Room Rearr & VirtualHome & ALFRED  & Watch-And-Help \\
    \midrule
    overall procedural plan & \textcolor{blue}{\cmark} & \textcolor{orange}{\xmark} & \textcolor{blue}{\cmark} & \textcolor{orange}{\xmark} & \textcolor{orange}{\xmark} & \textcolor{orange}{\xmark} & \textcolor{orange}{\xmark} & \textcolor{orange}{\xmark} \\
    activity reflects human behavior & \textcolor{blue}{\cmark}      &  \textcolor{blue}{\cmark}  & \textcolor{blue}{\cmark}         & \textcolor{blue}{\cmark} & \textcolor{orange}{\xmark}                 & \textcolor{orange}{\xmark}   & \textcolor{orange}{\xmark} & \textcolor{orange}{\xmark}  \\
    counterfactual                                                & \textcolor{blue}{\cmark}      &   \textcolor{orange}{\xmark}         & \textcolor{blue}{\cmark}         &  \textcolor{orange}{\xmark}   & \textcolor{orange}{\xmark}  & \textcolor{orange}{\xmark} & \textcolor{orange}{\xmark} & \textcolor{orange}{\xmark} \\
    multi-modality & \textcolor{blue}{\cmark} & \textcolor{blue}{\cmark} & \textcolor{orange}{\xmark} & \textcolor{orange}{\xmark} &  \textcolor{orange}{\xmark} & \textcolor{orange}{\xmark} & \textcolor{orange}{\xmark} & \textcolor{orange}{\xmark} \\
    activities   & 153     & 3,269  & - & 100  & 1    & 549    & 7  & 5  \\
    scenes/rooms       & 15/100   & 419 & - & 15/100      & -/120              & 6/24        & -/120   & 7/29           \\
    object categories    & 391      & 558 & - & 391         & 118                & 308         & 84      & 117            \\
    source  & iGibson2,ChatGPT & GPT-4 & GPT-3 & iGibson2    & AI2THOR            & Virtualhome & AI2THOR & Virtualhome    \\
    %task planning and/or control                                  & TP       & TP+C        & TP+C        & TP                 & TP          & TP      & TP \\
    \bottomrule
    \end{tabular}
    }
\caption{Comparison of ActPlan-1K to previous benchmarks related to household activities. Multi-modality means if the dataset provides both textual and visual task descriptions.}
\vskip-1.5em
\label{table:compare}
\end{table*}

To collect the multi-modal household activity dataset, we adopt iGibson2 \cite{2021iGibson} to generate environment scenes where the activities happen. iGibson2 is a household activity simulator providing customized activity definition and corresponding scene sampling. For
activities defined in Behavior-100 \cite{2021BEHAVIOR}, we translate the activity definition in BDDL symbolic form to natural language task description. Environment images are sampled in the simulator by loading the BDDL activity definition. Both the natural language task description and environment images are collected as the multi-modal instance. 

To further evaluate the counterfactual planning ability under constrained situations in real-world applications \cite{brahman2023plasma}, we design counterfactual activities based on normal activities defined in Behavior-100. Specifically, we request ChatGPT to generate situated conditions on the normal activities. Human annotators select the the situated conditions which can be defined and sampled in the household simulator. Counterfactual activity instances are then collected in the same way as normal activities.

%While solving realistic problems often involve thinking a best solution for various conditions even for a same goal, lack of benchmarks evaluating how these agents can solve the exceptions in physical world. Complex reasoning ability is under explored such as condition manipulations \cite{kim2022cosim}. We also defines the complex tasks for situational reasoning under physical conditions. 

Our benchmark covers 153 activities, 1,187 activity instances in total. Each activity instance contains 2\~{}5 images. The gold procedure length is 23.95 and 31.93 for normal and situated activities respectively. We evaluate typical VLMs (i.e., Claude \cite{anthropic2024claude}, Gemini-Pro \cite{team2023gemini}, and GPT-4V \cite{achiam2023gpt}) that are capable of conducting the task. Evaluation results on human score (correctness and commonsense satisfaction) show that current VLMs are still struggling to generate procedural plans of high-quality. To enhance future study with the proposed benchmark, we also propose automatic evaluation metrics including least common subsequence (LCS) score and finetuned BLEURT \cite{sellam2020bleurt} accuracy score. The dataset is available\footnote{https://github.com/HKUST-KnowComp/ActPlan-1K}.

\section{Related Work}

\subsection{Household Embodied AI benchmarks}
A substantial body of work constructs embodied environments or simulators, such as AI2THOR \cite{kolve2017ai2}, Gibson \cite{xia2018gibson}, iGibson \cite{shen2021igibson, 2021iGibson}, VirtualHome \cite{puig2018virtualhome}, AIHabitat \cite{savva2019habitat}. Based on them, VirtualHome \cite{puig2018virtualhome} crowdsources commonsense information about typical activities in people's homes and collect programs that formalize the activity in natural language. Watch-And-Help \cite{puig2020watch} focuses on social perception and human-AI collaboration. ALFREAD \cite{2019ALFRED} consists of expert demonstrations in interactive visual environments for 25k natural language directives from AI2THOR2.0 \cite{kolve2017ai2}. Behavior-100 \cite{2021BEHAVIOR} defines the various household activities to enrich physical and semantic properties with iGibson2.0 \cite{2021iGibson}.

%SAPIEN \cite{xiang2020sapien}, and ThreeDWorld \cite{gan2021threedworld}. 
%Various benchmarks are built based on the simulators evaluating the agents' ability over different dimensions. covering physical-aware navigation and interaction \cite{puig2018virtualhome, 2019ALFRED}, object states \cite{2021iGibson} and activities diversity \cite{2021BEHAVIOR, pmlr-v205-li23a}, tool usage skills \cite{2021ManiSkill, gan2022threedworld, gu2023maniskill2}, multi-agent interaction \cite{puig2020watch, 2021TEACh}. 

\begin{figure*}[t]
\centering
\includegraphics[scale=0.5, trim={0.5cm 0 0.5cm 0}]{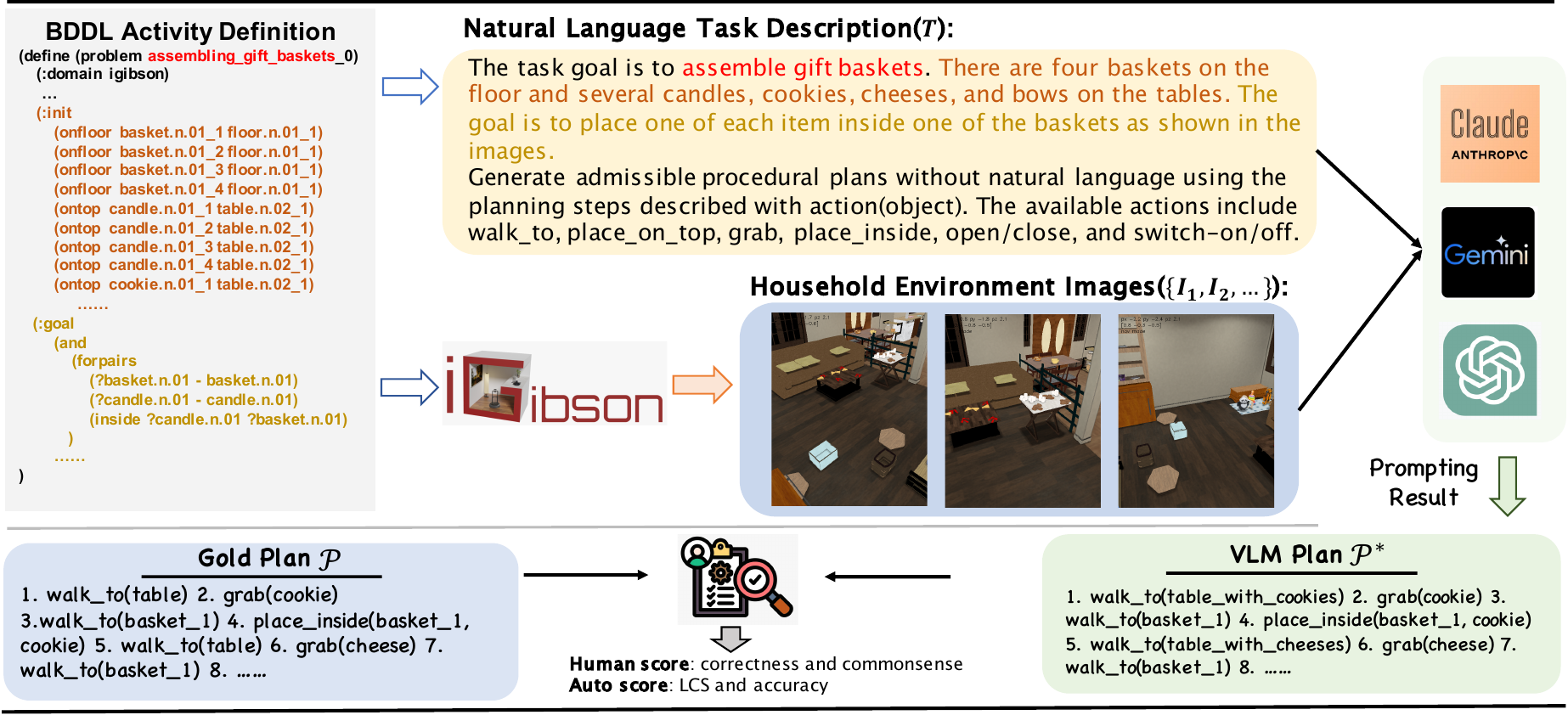}
\caption{Overview of ActPlan-1K dataset collection and evaluation. The BDDL activity definition is converted into natural language description. Environment images are sampled in the simulator after loading the BDDL definition. After prompting VLMs, VLM plan is evaluated by both human score and automatic metrics compared to gold plan. }
%\vskip-1.5em
\label{fig:overview}
\end{figure*}

Recent researches leverage the emergent reasoning capabilities of LLM and collect datasets through few-shot prompting. COPLAN \cite{brahman2023plasma} iteratively expands the planning dataset in natural language by prompting GPT-3 with seed examples. EgoPlan-Bench \cite{chen2023egoplan} is an instruction-tuning dataset from videos with human-object interactions, evaluating the VLMs' ability in predicting next-step actions in question-answering form. Our ActPlan-1k collects multimodal inputs based on both the simulator iGibson2 and ChatGPT. Comparison of our work to previous works is listed in Table \ref{table:compare}.

%ThreeDWorld Transport \cite{gan2021threedworld} focuses on physics-aware interaction and navigation by exploring the house, searching and transporting objects with two 9-DOF articulated arms. 

\subsection{Planning with LLMs and VLMs}
The use of large language models or vision language models in embodied AI tasks have become an increasingly popular research topic, by making procedural plans for high-level tasks with actionable and commonsense knowledge \cite{ahn2022can, raman:neurips22fmdm, li2022pre, Huang2022LanguageMA, du2023guiding, song2023llmplanner}.

LM-Nav \cite{shah2022lmnav} adopts LLMs to parse free-form instructions into landmarks for a vision-language model to infer a joint probability over the landmarks and images. GPT-3 and Codex can produce plausible action plans with example of task description and its associated action sequences \cite{Huang2022LanguageMA}. LLM-Planner \cite{song2023llmplanner} design strategies for LLMs to generate high-level plans dynamically to accomplish final goal in interactive manner with few-shot prompting. The multi-modal LLM PaLM-E \cite{driess2023palm} takes in images, state estimates, or other sensor modalities to generate executable sequence plans. In Text2Motion \cite{lin2023text2motion}, a library of learned skills are adopted to interface with an LLM for generating feasible plans prior to execution. AutoRT \cite{ahn2024autort} utilizes both the vision-language models (VLMs) and large language models (LLMs) to generate and determine executable high-level objectives. While LLMs have been exploited in the embodied AI tasks mostly with textual description, we target at benchmarking the planning ability of VLMs with both textual and visual descriptions. 

\section{ActPlan-1K Benchmark}
ActPlan-1K contains multi-modal planning instances both of normal activities and counterfactual activities, with same instance form in Section \ref{sec: problem}. The instance collection process of them is also same with BDDL definition and illustrated in Section \ref{sec: actplan-1k}. Details of constructing counterfactual activities are introduced in Section \ref{sec:counterfactual}. The benchmark statistics is presented in Section \ref{sec: statistics}.
\subsection{Problem Definition}
\label{sec: problem}
Given household environment $\mathcal{E}$, there are manipulable objects set $\mathcal{O}$. For each household activity $\mathcal{T}$, an VLM agent $\mathcal{A}$ takes in the task description $T$ and environment images $\{I_1, I_2, ...\}$ as input, generates procedural plan $\mathcal{P^*}$ that can accomplish the task. The household environments have multiple interior spaces therefore there are multiple images to ensure that necessary spatial information is provided. The overall framework is presented in Figure \ref{fig:overview}. 

The gold procedural plan $\mathcal{P}$ consists of multi-step action sequences which are conducted on the objects. To facilitate the evaluation of high-level plans in real-world scenarios, we design symbolic action plans followed the design in VirtualHome \cite{puig2018virtualhome}. Specifically, we use 8 main actions, namely, \textit{walk\_to}, \textit{place\_on\_top}, \textit{grab}, \textit{place\_inside}, \textit{open/close}, \textit{switch-on/off}. Additional actions such as \textit{wipe} and \textit{pour} are permitted if they satisfy real-life scenario commonsense. 

\subsection{ActPlan-1K Instance Collection}
\label{sec: actplan-1k}
To construct multi-modal input for describing the household activity, we use an household activity simulator iGibson2 \cite{2021iGibson}. The iGibson2 simulator\footnote{https://svl.stanford.edu/igibson/} provides visual components for image collection, including renderer and viewer. In the simulator, 15 household environments with 342 types of objects are provided for activity scene initialization. 

iGibson2 takes in BDDL definitions in BDDL and loads household environment according to the BDDL activity definition. BDDL provides predicate-logic descriptions for object-centric activity design. New activities with customized initialization of objects and scene can be defined with BDDL language\footnote{https://behavior.stanford.edu/activity-annotation} and then loaded in the simulator\footnote{https://stanfordvl.github.io/iGibson/sampling.html}. 

We use seed BDDL activities from Behavior-100 \cite{2021BEHAVIOR} and load them in selected household environment $\mathcal{E}$ in iGibson2. In viewer mode, we record the touring video around the indoor scenes. Images covering the main information in BDDL activity definition are selected as visual information $\{I_1, I_2, ...\}$. Generally, 2 to 5 images are selected to cover the main contents. As for natural language description, we transform the BDDL activity description into natural language form $T$ by prompting ChatGPT with few-shot examples. As for gold plan $\mathcal{P}$, we ask human annotators to write in action sequences over object subset from $\mathcal{O}$. The actions are mostly from pre-defined 8 actions and the objects are mostly from the activity definition. Actions can also be conducted on the objects in the images which are not in the definition if necessary. An example of multi-modal input and output is presented in Figure \ref{fig:overview}. 

\subsection{Counterfactual Activity}
\label{sec:counterfactual}
Counterfactual planning necessitates agents to engage in reasoning about alternative constrained scenarios and formulate corresponding plans \cite{brahman2023plasma}. The ability to engage in counterfactual planning holds significant importance for VLM agents, as constrained situations frequently arise in real-world applications. Enhancing this capability enables agents to effectively navigate and respond to complex and dynamic environments.

We define multi-modal counterfactual activities by adjusting the task descriptions and gold plans from normal activities in Behavior-100. Behavior-100 activity definitions are common ones and require normal procedures to accomplish the goal. To acquire counterfactual activities based on normal activities in Behavior-100, there are three steps as follows:
\begin{itemize}
    \item Step 1: Request ChatGPT for common procedures to accomplish the given activities, and request for what might happen during the procedure of common activities;
    \item Step 2: Human selection of the circumstances that are commonsense plausible, and suitable for BDDL definition;
    \item Step 3: Load the new counterfactual BDDL activity definition in iGibson2 simulator and collect the visual inputs. 
\end{itemize}
The detailed pipeline of construction is presented in Appendix \ref{Appendix:1}. An prompting example for Step 1 is presented in Figure \ref{fig:append 1}.

To ensure the counterfactual activity BDDL definition can be loaded for image collection, human annotation is required in Step 2. The selected counterfactual circumstances should also be commonsense plausible, which means reasonable and acceptable based on human understanding and knowledge of the world. 

\begin{figure}[t]
\centering
\includegraphics[scale=0.6, trim={0.5cm 0 0.5cm 0}]{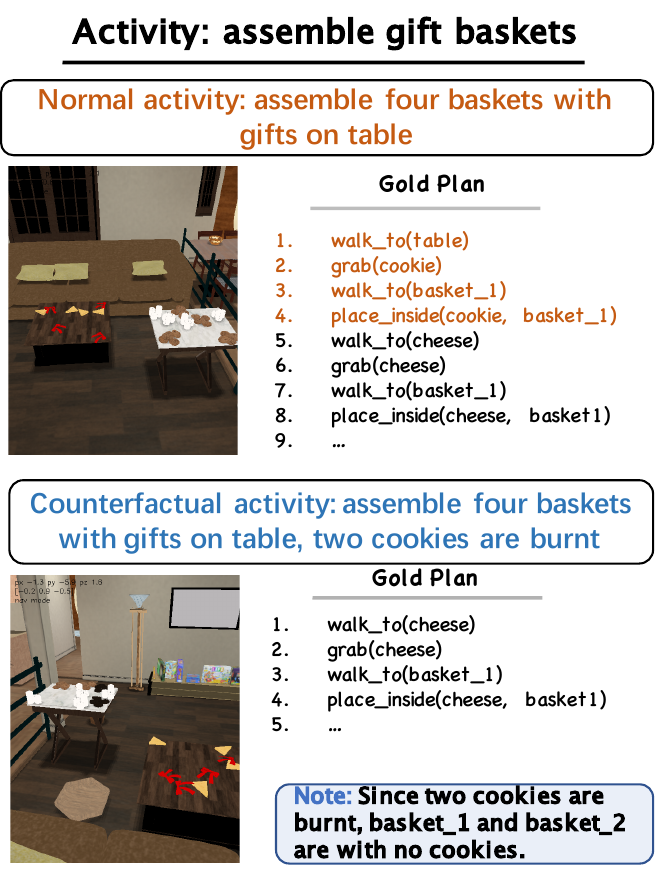}
\caption{Example of normal activity and counterfactual activity. The gold plans are different because of the counterfactual situation.}
\vskip-1.5em
\label{fig:counterfactual}
\end{figure} 

After loading the counterfactual definition in the simulator, environment images are collected in Step 3. An detailed example of normal activity and counterfactual activity is in Figure \ref{fig:counterfactual}. In the example, the task inputs and gold plans are different between the two activities. Gold procedural plans for ``assemble gift baskets'' will distribute each cookie into each basket. However when two cookies are burnt, they will not be suitable as gifts so they should not be assembled in the gift baskets. Therefore, this situation will lead to a different procedural plan.

\subsection{Activity Statistics}
\label{sec: statistics}
The counterfactual activities can be broadly classified into three categories:\\
\noindent \textbf{Object Property}. Under this situation, the plan need to be adjusted in order to consider the special properties of object to adjust the plans. An example is shown in Figure \ref{fig:counterfactual}. In the example, burnt cookies are not suitable as gifts, so two baskets (basket\_1 and basket\_2) will not be assembled cookies. The property of cookie (burnt and unburnt) is considered to adjust the plan.\\
\noindent \textbf{Object Function}. The functionalists of different objects are considered and designed to solve a special problem. An example is shown in Appendix \ref{Appendix:2} Figure \ref{fig:counterfactual2}. \\
\noindent \textbf{Event Causality}. To accomplish the activity goal, additional steps are required to handle unexpected events in the middle of the normal procedures. An example is shown in Appendix \ref{Appendix:2} Figure \ref{fig:counterfactual3}. 

\begin{figure}[t]
\centering
\includegraphics[scale=0.6, trim={0 0 1cm 0}]{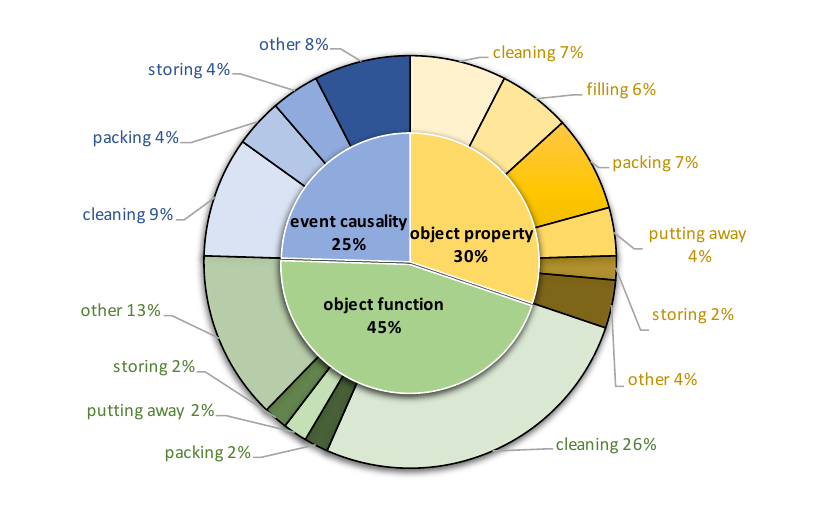}
\caption{Distribution of counterfactual activities.}
\vskip-0.5em
\label{fig:situated}
\end{figure}

In detail, we group activities in each category them by activity types: \textit{cleaning}, \textit{filling}, \textit{packing}, \textit{putting away}, \textit{storing}, and \textit{other}. The distribution is presented in Figure \ref{fig:situated}.

For both normal activities and counterfactual activities, each activity is loaded in three randomly selected household environments in the simulator. To diversify the visual content, each activity in each household environment is sampled up to three times in each household environment. The overall instance distribution is listed in Table \ref{tab:dist}. In general, the counterfactual activities have longer plan sequences than normal activities since additional steps are required to handle counterfactual situations.

\begin{table}[]
    \centering
    \begin{tabular}{c|c|c|c}
    \toprule
     Type    &  Num & Instances & Avg len \\
    \midrule
     Normal  & 100 & 825 & 23.95 \\
    \midrule
     Counterfactual & 53 & 362 & 31.93 \\
    \bottomrule
    \end{tabular}
    \caption{Distribution of activity instances. Avg len represents the average length of plan sequences.}
    \vskip-1.0em
    \label{tab:dist}
\end{table}

\section{Evaluation Metrics}

Evaluation metrics over VLM generated procedural plan $\mathcal{P}^*$ and gold plan $\mathcal{P}$ have two types: human evaluation and automatic evaluation.

\subsection{Human Evaluation}
The activity plans in ActPlan-1K are defined to have complex and long procedural steps, and can not be directly evaluated in the simulator. Therefore, we ask human annotators to annotate the correctness and commonsense satisfaction of $\mathcal{P}^*$. \\
\textbf{Correctness}. The correctness is 1 if the entire plan can achieve the final activity goal. The plan steps does not strictly follows the step orders of the gold plan. Otherwise the correctness is 0. \\
\noindent \textbf{Commonsense Satisfaction}. The commonsense satisfaction is 1 if every procedural step is commonsense plausible. The final goal is not necessarily achieved. If one of the procedural step does not satisfy commonsense constraints, the commonsense satisfaction is 0.\\

%Unlike VirtualHome \cite{puig2018virtualhome} where executability in scenes can be checked in the simulator,

\subsection{Automatic Evaluation}
Automatic evaluation is conducted based on the N-gram based metric: longest common subsequence (LCS),  and BLEURT-based accuracy score.
\subsubsection{LCS}
Similar to \cite{puig2018virtualhome}, we calculate the LCS between two plans $\mathcal{P}^*$ and $\mathcal{P}$. The normalized LCS score is also calculated by normalizing LCS length with the maximum length of the two plans. To judge if two steps are semantically equivalent, we use a sentence transformer \textit{all-MiniLM-L6-v2}\footnote{https://huggingface.co/sentence-transformers/all-MiniLM-L6-v2} to compare the similarity of two plan steps. Two plan steps are semantic similarly if the score is greater than 0.8.

\begin{table*}[t]
\resizebox{\textwidth}{!}{
\begin{tabular}{lcccccc}
\toprule
\multirow{2}{*}{Model} & \multicolumn{2}{c}{Normal activity}                           & \multicolumn{2}{c}{Counterfactual activity}                         & \multicolumn{2}{c}{Total}      \\ \cline{2-7}
                       & \multicolumn{1}{c}{Correctness} & \multicolumn{1}{c}{Commonsense} & \multicolumn{1}{c}{Correctness} & \multicolumn{1}{c}{Commonsense} & \multicolumn{1}{c}{Correctness} & \multicolumn{1}{c}{Commonsense} \\
\midrule
%Human & 100.0 & 100.0 & 100.0 & 100.0 & 100.0 & 100.0 \\
%\midrule
Claude-3-haiku  & 8.0 & 13.8 & 3.9 & 9.4 & 6.7 & 12.4 \\
Claude-3-sonnet & 22.2 & 34.9 & 10.2 & 29.9 & 18.4 & 25.9 \\
Gimini-Pro-1.5 & 32.0 & 44.0 & 22.8 & 41.7 & 29.1 & 43.3 \\
GPT-4V  &  29.1 & 39.6 &  9.4 & 25.2 &  22.3  &  39.6  \\
\bottomrule
\end{tabular}
}
\caption{Correctness score (\%) and commonsense satisfaction score (\%) of VLMs on ActPlan-1K.}
\vskip-0.5em
\label{ret:1}
\end{table*}

\subsubsection{Accuracy}
By converting the plan sequences into sentence pairs, we design a classification task to automatically evaluate the quality of generated plan. We follow the task-specific finetuning schema in BLEURT \cite{sellam2020bleurt}. The BLEURT is a learned evaluation metric based on BERT \cite{kenton2019bert} that can model human judgements with small task-specific data on text generation tasks. We finetune BLEURT models on ActPlan-1K plans to build automatic evaluation on generated plans.

Given sentence pair $\mathbf{x}$ and $\mathbf{\tilde{x}}$, $\mathbf{x}$ is a sentence for gold plan $\mathcal{P}$ and $\mathbf{\tilde{x}}$ is a sentence for generated plan $\mathcal{P}^*$. $\{(\mathbf{x}, \mathbf{\tilde{x}}, y_i)\}^N_{n=1}$ is a training dataset of size $N$, where $y_i$ is a classification label that indicates if $\mathbf{\tilde{x}}$ is semantically equivalent to $\mathbf{x}$. 

$\mathbf{x}$ and $\mathbf{\tilde{x}}$ are as input to a BLEURT transformer, which has been finetuned on large amount of synthetic data to model human rating. A sequence of contextualized vectors are returned as:
\begin{equation}
    \mathbf{v}_{[CLS]}, \mathbf{v}_1, ... = {\rm BLEURT}([\mathbf{x}, \mathbf{\tilde{x}]}),
\end{equation}
where $[\mathbf{x}, \mathbf{\tilde{x}}]$  stands for the concatenation of $\mathbf{x}$ and $\mathbf{\tilde{x}}$ concatenated with a $[SEP]$ token. $\mathbf{v}_{[CLS]}$ is the representation for the special $[CLS]$ token. With a linear layer on top of the $[CLS]$ vector, the classification label is predicted as:
\begin{equation}
    \tilde{y} = \mathbf{W}\mathbf{v}_{[CLS]} + \mathbf{b},
\end{equation}

With a few thousand examples as supervised data, a supervised classification loss:
\begin{equation}
    loss = - \frac{1}{N}\sum^{N}_{i=1}[y_i\log(p_i)+(1-y_i)\log(1-p_i)],
\end{equation}
where $y_i \in [0,1]$ and $p_i$ is the softmax probability.

To finetune the model, we construct synthetic training sequences from gold plans and generated plans from VLMs (i.e., Gemini-Pro-1.5 \cite{team2023gemini} and GPT-4V \cite{achiam2023gpt}). Specifically, the plan sequences are split into train/val/test with 60/20/20 normal activities, and the corresponding counterfactual activities. For each activity instance, more than two gold sequences are labeled. For generated plans, the correctness scores are human labeled. 

The gold plans and correct plans are mixed to form positive sentence pairs. To form negative sentence pairs with the gold plan, wrong plan from VLMs or randomly shuffled gold plan with same activity is selected. The synthetic data are used to finetune \texttt{BLEURT}(based on BERT-large) and \texttt{BLEURT-base}(based on BERT-base) with a classification layer on top of them. Learning rate is 5e-6 for \texttt{BLEURT} and 1e-6 for \texttt{BLEURT-base}. Batch size is set as 12. Each model is finetuned for 70 epochs. The model achieves best score on val set is selected as the best model. \\

\section{Experiment Result}

\subsection{Experiment Setup}
By adopting VLMs as agent $\mathcal{A}$, the input are task description in natural language form $T$ and sampled environment images $\{I_1, I_2, ...\}$, forming as multi-modal input: 
\begin{equation}
    \mathcal{P^*} = \text{VLM}([T, I_1, I_2, ...])
\end{equation}
The VLMs are Claude-3-haiku and Cluade-3-sonnet\footnote{https://www.anthropic.com/news/claude-3-family}, Gemini-Pro-1.5 \cite{team2023gemini}, and GPT-4V \cite{achiam2023gpt}. The decoding temperature is  0 for all the models in the evaluation. Please see Appendix \ref{Appendix:3} for additional details.

\begin{figure}[t]
\centering
\includegraphics[scale=0.5, trim={0.5cm 0 0.5cm 1cm}]{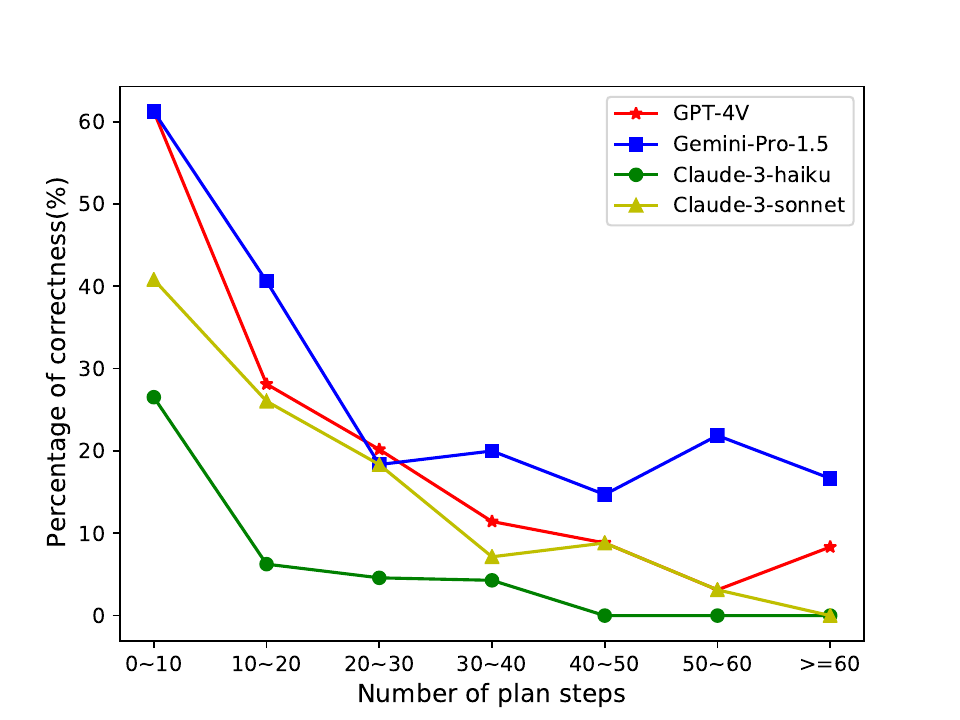}
\caption{Correctness (\%) with plan sequence length.}
\vskip-1.0em
\label{fig:correctness}
\end{figure}

\subsection{Correctness and Commonsense Satisfaction}
Human evaluation results on VLM generated plans are shown in Table \ref{ret:1}. On both the correctness and commonsense metrics, the scores are low and this shows that procedural planning is still challenging for VLMs.

\textbf{Counterfactual activities are harder to accomplish}. VLMs achieves much higher scores on normal activities than counterfactual activities. It demonstrates that counterfactual reasoning ability over embodied household activities still requires enhancement. Comparing correctness and commonsense satisfaction, VLMs are more capable to provide the commonsense knowledge required for both type of activities, while not as good as designing accurate procedural steps. 

\textbf{Gemini-Pro achieves better performance than other VLMs}. Comparing the VLMs, we can find that Gemini-Pro achieves better performances on all the human evaluation metrics. Compared to GPT-4V, the strength mainly lies on the ability to accomplish counterfactual activities. Combined with Figure \ref{fig:correctness}, we can find that the strength also lies on plans with longer sequence plans.

\textbf{Correctness is degrading with sequence length increases}. As presented in Figure \ref{fig:correctness}. The correctness of sequence length drops greatly with the sequence length increases. Long-sequence plan is hard as it needs to take consideration of more objects and events. The phenomenon is same for all the four VLMs.

\subsection{Automatic Evaluation}

\subsubsection{LCS}
Evaluation results of the LCS and Norm LCS scores are shown in Table \ref{ret:2}. Results show Gemini-Pro-1.5 achieves best performances on LCS and Norm.LCS among the VLMs. About half of the sequences from GPT-4V match the correct ones. Claude-3-haiku and Claude-3-sonnet achieves much lower scores than GPT-4V and Gemini-Pro-1.5, which is consistent with the correctness and commonsense score in Table \ref{ret:1}. LCS and Norm LCS scores provide an easy and approximate evaluation metric for the generated plans.

\begin{table}[]
\centering
\begin{tabular}{lcc}
\toprule
Model & LCS & Norm. LCS  \\
\midrule
Claude-3-haiku     &      8.62     & 0.38  \\
Claude-3-sonnet    &     11.87     & 0.48 \\
%Gemini-Pro-1.5        &    13.29   &  0.50 \\
Gemini-Pro-1.5        &    18.00   &  0.67 \\
GPT-4V            &    14.23   &     0.55    \\
\bottomrule
\end{tabular}
\caption{LCS and Norm LCS Score (\%) of VLMs on ActPlan-1K.}
\label{ret:2}
%\vskip-1.5em
\end{table}

\subsection{Accuracy}

The results of accuracy on val and test splits from Gemini-Pro and GPT-4V for finetuning BLEURT models are presented in Table \ref{ret:4} separately. On both \texttt{BLEURT-base} and \texttt{BLEURT}, the accuracy scores are all above 60\%, achieving as high as 81.5\%. This shows that automatic classification task has promising results in modeling human evaluation on the correctness metric.

Since the correctness labels are significantly imbalanced as many are 0, we further calculate the Matthews Correlation Coefficient(MCC) between the original correctness labels and predicted classification results on val split. The MCC score measures the correlation of human label and automatic predicted labels. Results in Table 4 show that on \texttt{BLEURT}, the MCC achieves 0.321 and 0.367 on Gemini-Pro and GPT-4V, close to moderate standard. To further improve the correlation, increasing the training data amount would help make the evaluation results more close to human ratings.

%\textbf{MCC is a compatible evaluation metric to accuracy}. The gold labels are significantly imbalanced. Simply improve the accuracy would lead to the model stressing on output negative labels on current test set. With MCC score, we can find that the automatic metrics is correlating to human labeling. The MCC score achieves 0.321 and 0.367 on Gemini-Pro and GPT-4V, greatly improves the correlation between automatic accuracy and human evaluated correctness. 

\texttt{BLEURT} achieves better performance than \texttt{BLEURT-base} due to its larger model size. The performance gain is obvious on MCC and accuracy score for both of the VLMs. Results with \texttt{BLEURT} would make the automatic evaluation on accuracy more robust to use.

\begin{table}[]
\centering
\resizebox{0.5\textwidth}{!}{
\begin{tabular}{lcccc}
\toprule
Model    & VLM       & MCC &  Val(Acc)     & Test(Acc) \\ 
\midrule
\multirow{2}{*}{\texttt{BLEURT-base}} & Gemini-Pro-1.5      & 0.258 & 65.4 &  65.4  \\ 
  & GPT-4V          & 0.258 & 79.0 &  70.4  \\ 
\midrule
\multirow{2}{*}{\texttt{BLEURT}} & Gemini-Pro-1.5      & 0.321 & 75.3 & 80.2 \\
 & GPT-4V          & 0.367 & 81.5 & 76.5 \\
\bottomrule
\end{tabular}
}
\caption{Accuracy (\%) and MCC of VLMs on ActPlan-1K val and test splits.}
\label{ret:4}
%\vskip-0.5em
\end{table}

\begin{figure*}[t]
\centering
\includegraphics[scale=0.45, trim={0.5cm 0 0.5cm 0.5cm}]{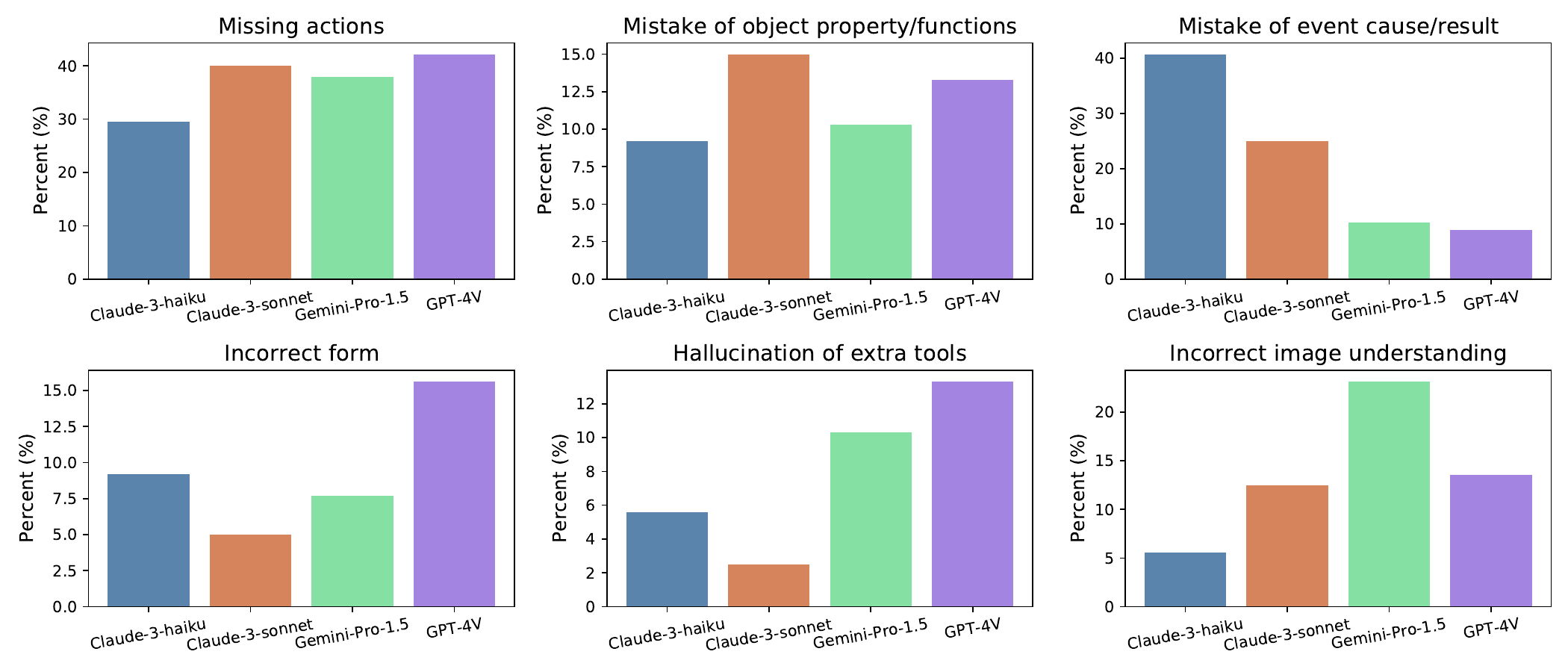}
\caption{Error distribution (\%) of VLMs on six types of errors.}
%\vskip-1.5em
\label{fig:error type}
\end{figure*}

%\begin{table}[]
%\centering
%\resizebox{0.5\textwidth}{!}{
%\begin{tabular}{lccc}
%\toprule
%  Gemini-Pro-1.5    & Correctness & Commonsense & Avg Len \\
%\midrule
%w image  &     36.7      &      43.3    & 26.93  \\
%\midrule
%w/o image  &      26.7       &     35.0   & 52.63   \\
%\bottomrule
%\end{tabular}
%}
%\caption{Ablation study results of images as prompting input to Gemini-Pro-1.5.}
%\label{ret:abla}
%\vskip-1.5em
%\end{table}

\begin{table}[]
\centering
\resizebox{0.5\textwidth}{!}{
\begin{tabular}{lccc}
\toprule
Model(w or w/o image) & Correctnetss & Commonsense & Avg.len \\
\midrule
ChatGPT(w/o)   & 13.3         & 25.0        & 18.76   \\
\midrule
Gemini-Pro-1.5(w/o)   & 26.7         & 35.0        & 52.63 \\ 
\midrule
Gemini-Pro-1.5(w)  & 36.7         & 43.3        & 26.93  \\
\bottomrule
\end{tabular}
}
\caption{Ablation study results of no images as prompting input to ChatGPT and Gemini-Pro-1.5.}
\label{ret:abla}
\end{table}

\subsection{Ablation Study}
To see if the images are necessary or provide key information for procedural planning, we prompt ChatGPT and Gemini-pro with text description only. Results of 40 normal activities with corresponding 20 counterfactual activities are presented in Table \ref{ret:abla}. Further detailed results in normal and counterfactual ones are in Appendix \ref{appendix:abla}. Results show that without images, the correctness or commonsense quality drops while with longer steps. The reasons are:

\textbf{ActPlan-1K is challenging for LLM with no visual information considered}. ChatGPT generates much shorter responses than Gemini-Pro-1.5 on average, and the correctness and commonsense scores are much lower than Gemini-Pro-1.5.

\textbf{Inaccurate understanding of objects and events}. The textual input is lacking of background information, leading to inaccurate plan generation. For example, there is usually sink in bathroom and kitchen which can be used as water source and wash items. If not specifically pointed out, Gemini-pro would generate plans without these utilities and lead to plan failure.

\textbf{Hallucination leads to longer steps}. Since there is lack of visual input of the spatial information, Gemini-pro generates repetitive steps (i.e., \textit{walk\_to(kitchen), walk\_to(table)} before grab items in kitchen), or hallucinates on the specific states of objects that need to be handled. 

\subsection{Error Analysis}

We conduct error analysis by sampling 40 activity plans (including both normal and counterfactual activities) from Claude-3-haiku, Claude-3-sonnet, Gemini-Pro-1.5 and GPT-4V. The results are presented in Figure \ref{fig:error type}. Details of example instances are presented in Appendix \ref{Appendix:4}.

In both normal and counterfactual activities, the error analysis are broadly categorized as:
\begin{itemize}
    \item \textbf{Missing actions}. Between consecutive sub-sequences, there are lack of actions to complete. For example, after grabbing food in refrigerator, there is lack of ``walk\_to(sink)'' before placing the food into sink.
    \item \textbf{Mistake of object property/function}. Mistakes caused by misunderstanding the properties or functions of the objects, which are mostly from situated circumstances, such as failure to recognize that burnt cookies is not suitable as gifts, salt can help ice melting instead of vinegar, etc.
    \item \textbf{Mistake of event cause/effect}. Mistakes caused by misunderstanding the normal procedures required for an activity. For example, by adopting soap and rag to clean shoes, soap and rag should be soaked before cleaning, not directly to do cleaning in dry state.
    \item \textbf{Incorrect form}. Mistakes caused by generating sub-sequences not in the required ``action(object)'' form. 
    \item \textbf{Hallucination of extra tools}. Error in generating sub-sequences by utilizing extra tools which are not shown in the images or task descriptions.
    \item \textbf{Incorrect image understanding}. Error in understanding accurate contents in the images, such as the distance between refrigerator, number of clean plates in the images. 
\end{itemize}

From Figure \ref{fig:error type}, VLMs have different error distributions on the various error types. Missing actions is the mostly occurred error and mistake of object property/functions is a common occurred error for all VLMs. 
\section{Conclusion}
We present ActPlan-1K, a multi-modal counterfactual household planning dataset, to evaluate both the procedural planning and counterfactual planning ability of vision language models. Our dataset constructed based on ChatGPT and household simulators consists of 1.2k instances covering normal and counterfactual activities. Evaluation on typical VLMs shows that ActPlan-1K is still challenging for the models. 

\section*{Limitations}
The images collected from iGibson2 simulator are of low-resolution which may have influence on image understanding of VLMs. The limited types of object states and scene sampling constraints in the simulator restrict the definition of counterfactual activities. While VLMs as agents for embodied AI tasks is an important and promising research direction, we believe that future development of household simulators will benefit better construction of evaluation benchmarks of our kind.

\section*{Acknowledgements}
The authors of this paper were supported by the NSFC Fund (U20B2053) from the NSFC of China, the RIF (R6020-19 and R6021-20) and the GRF (16211520 and 16205322) from RGC of Hong Kong.

%\section{Experiment}
%\subsection{Benchmark}
%\noindent \textbf{TruthfulQA}.

%Commonsense reasoning related tasks: TB-Dense, MATRES, TDDMan, COPA, e-CARE, HeadlineCause, PDTB Explicit Discourse Relation, PDTB implicit Discourse Relation, DiscoGen All Subsets, DiscoGen Europarl Subset, DiscoGen Novel Subset, DiscoGen, Wikipedia Subset.

\bibliography{anthology,custom}

\begin{thebibliography}{31}
\expandafter\ifx\csname natexlab\endcsname\relax\def\natexlab#1{#1}\fi

\bibitem[{Achiam et~al.(2023)Achiam, Adler, Agarwal, Ahmad, Akkaya, Aleman,
  Almeida, Altenschmidt, Altman, Anadkat et~al.}]{achiam2023gpt}
Josh Achiam, Steven Adler, Sandhini Agarwal, Lama Ahmad, Ilge Akkaya,
  Florencia~Leoni Aleman, Diogo Almeida, Janko Altenschmidt, Sam Altman,
  Shyamal Anadkat, et~al. 2023.
\newblock Gpt-4 technical report.
\newblock \emph{arXiv preprint arXiv:2303.08774}.

\bibitem[{Ahn et~al.(2022)Ahn, Brohan, Brown, Chebotar, Cortes, David, Finn,
  Fu, Gopalakrishnan, Hausman et~al.}]{ahn2022can}
Michael Ahn, Anthony Brohan, Noah Brown, Yevgen Chebotar, Omar Cortes, Byron
  David, Chelsea Finn, Chuyuan Fu, Keerthana Gopalakrishnan, Karol Hausman,
  et~al. 2022.
\newblock Do as i can, not as i say: Grounding language in robotic affordances.
\newblock \emph{arXiv preprint arXiv:2204.01691}.

\bibitem[{Ahn et~al.(2024)Ahn, Dwibedi, Finn, Arenas, Gopalakrishnan, Hausman,
  Ichter, Irpan, Joshi, Julian et~al.}]{ahn2024autort}
Michael Ahn, Debidatta Dwibedi, Chelsea Finn, Montse~Gonzalez Arenas, Keerthana
  Gopalakrishnan, Karol Hausman, Brian Ichter, Alex Irpan, Nikhil Joshi, Ryan
  Julian, et~al. 2024.
\newblock Autort: Embodied foundation models for large scale orchestration of
  robotic agents.
\newblock \emph{arXiv preprint arXiv:2401.12963}.

\bibitem[{Anthropic(2024)}]{anthropic2024claude}
AI~Anthropic. 2024.
\newblock The claude 3 model family: Opus, sonnet, haiku.
\newblock \emph{Claude-3 Model Card}.

\bibitem[{Brahman et~al.(2023)Brahman, Bhagavatula, Pyatkin, Hwang, Li, Arai,
  Sanyal, Sakaguchi, Ren, and Choi}]{brahman2023plasma}
Faeze Brahman, Chandra Bhagavatula, Valentina Pyatkin, Jena~D Hwang,
  Xiang~Lorraine Li, Hirona~J Arai, Soumya Sanyal, Keisuke Sakaguchi, Xiang
  Ren, and Yejin Choi. 2023.
\newblock Plasma: Making small language models better procedural knowledge
  models for (counterfactual) planning.
\newblock \emph{arXiv preprint arXiv:2305.19472}.

\bibitem[{Chen et~al.(2023)Chen, Ge, Ge, Ding, Li, Wang, Xu, Shan, and
  Liu}]{chen2023egoplan}
Yi~Chen, Yuying Ge, Yixiao Ge, Mingyu Ding, Bohao Li, Rui Wang, Ruifeng Xu,
  Ying Shan, and Xihui Liu. 2023.
\newblock Egoplan-bench: Benchmarking egocentric embodied planning with
  multimodal large language models.
\newblock \emph{arXiv preprint arXiv:2312.06722}.

\bibitem[{Driess et~al.(2023)Driess, Xia, Sajjadi, Lynch, Chowdhery, Ichter,
  Wahid, Tompson, Vuong, Yu et~al.}]{driess2023palm}
Danny Driess, Fei Xia, Mehdi~SM Sajjadi, Corey Lynch, Aakanksha Chowdhery,
  Brian Ichter, Ayzaan Wahid, Jonathan Tompson, Quan Vuong, Tianhe Yu, et~al.
  2023.
\newblock Palm-e: An embodied multimodal language model.
\newblock \emph{arXiv preprint arXiv:2303.03378}.

\bibitem[{Du et~al.(2023)Du, Watkins, Wang, Colas, Darrell, Abbeel, Gupta, and
  Andreas}]{du2023guiding}
Yuqing Du, Olivia Watkins, Zihan Wang, C{\'e}dric Colas, Trevor Darrell, Pieter
  Abbeel, Abhishek Gupta, and Jacob Andreas. 2023.
\newblock Guiding pretraining in reinforcement learning with large language
  models.
\newblock \emph{arXiv preprint arXiv:2302.06692}.

\bibitem[{Huang et~al.(2023)Huang, Mees, Zeng, and Burgard}]{huang23vlmaps}
Chenguang Huang, Oier Mees, Andy Zeng, and Wolfram Burgard. 2023.
\newblock Visual language maps for robot navigation.
\newblock In \emph{Proceedings of the IEEE International Conference on Robotics
  and Automation (ICRA)}, London, UK.

\bibitem[{Huang et~al.(2022)Huang, Abbeel, Pathak, and
  Mordatch}]{Huang2022LanguageMA}
Wenlong Huang, P.~Abbeel, Deepak Pathak, and Igor Mordatch. 2022.
\newblock \href {https://api.semanticscholar.org/CorpusID:246035276} {Language
  models as zero-shot planners: Extracting actionable knowledge for embodied
  agents}.
\newblock \emph{ArXiv}, abs/2201.07207.

\bibitem[{Kenton and Toutanova(2019)}]{kenton2019bert}
Jacob Devlin Ming-Wei~Chang Kenton and Lee~Kristina Toutanova. 2019.
\newblock Bert: Pre-training of deep bidirectional transformers for language
  understanding.
\newblock In \emph{Proceedings of NAACL-HLT}, pages 4171--4186.

\bibitem[{Kolve et~al.(2017)Kolve, Mottaghi, Han, VanderBilt, Weihs, Herrasti,
  Deitke, Ehsani, Gordon, Zhu et~al.}]{kolve2017ai2}
Eric Kolve, Roozbeh Mottaghi, Winson Han, Eli VanderBilt, Luca Weihs, Alvaro
  Herrasti, Matt Deitke, Kiana Ehsani, Daniel Gordon, Yuke Zhu, et~al. 2017.
\newblock Ai2-thor: An interactive 3d environment for visual ai.
\newblock \emph{arXiv e-prints}, pages arXiv--1712.

\bibitem[{Li et~al.(2021)Li, Xia, Martín-Martín, Lingelbach, Srivastava,
  Shen, Vainio, Gokmen, Dharan, and Jain}]{2021iGibson}
Chengshu Li, Fei Xia, Roberto Martín-Martín, Michael Lingelbach, Sanjana
  Srivastava, Bokui Shen, Kent Vainio, Cem Gokmen, Gokul Dharan, and Tanish
  Jain. 2021.
\newblock igibson 2.0: Object-centric simulation for robot learning of everyday
  household tasks.

\bibitem[{Li et~al.(2022)Li, Puig, Paxton, Du, Wang, Fan, Chen, Huang,
  Aky{\"u}rek, Anandkumar et~al.}]{li2022pre}
Shuang Li, Xavier Puig, Chris Paxton, Yilun Du, Clinton Wang, Linxi Fan, Tao
  Chen, De-An Huang, Ekin Aky{\"u}rek, Anima Anandkumar, et~al. 2022.
\newblock Pre-trained language models for interactive decision-making.
\newblock \emph{Advances in Neural Information Processing Systems},
  35:31199--31212.

\bibitem[{Lin et~al.(2024)Lin, Yin, Ping, Molchanov, Shoeybi, and
  Han}]{lin2024vila}
Ji~Lin, Hongxu Yin, Wei Ping, Pavlo Molchanov, Mohammad Shoeybi, and Song Han.
  2024.
\newblock Vila: On pre-training for visual language models.
\newblock In \emph{Proceedings of the IEEE/CVF Conference on Computer Vision
  and Pattern Recognition}, pages 26689--26699.

\bibitem[{Lin et~al.(2023)Lin, Agia, Migimatsu, Pavone, and
  Bohg}]{lin2023text2motion}
Kevin Lin, Christopher Agia, Toki Migimatsu, Marco Pavone, and Jeannette Bohg.
  2023.
\newblock Text2motion: From natural language instructions to feasible plans.
\newblock \emph{Autonomous Robots}, 47(8):1345--1365.

\bibitem[{Puig et~al.(2018)Puig, Ra, Boben, Li, Wang, Fidler, and
  Torralba}]{puig2018virtualhome}
Xavier Puig, Kevin Ra, Marko Boben, Jiaman Li, Tingwu Wang, Sanja Fidler, and
  Antonio Torralba. 2018.
\newblock Virtualhome: Simulating household activities via programs.
\newblock In \emph{Proceedings of the IEEE Conference on Computer Vision and
  Pattern Recognition}, pages 8494--8502.

\bibitem[{Puig et~al.(2020)Puig, Shu, Li, Wang, Liao, Tenenbaum, Fidler, and
  Torralba}]{puig2020watch}
Xavier Puig, Tianmin Shu, Shuang Li, Zilin Wang, Yuan-Hong Liao, Joshua~B
  Tenenbaum, Sanja Fidler, and Antonio Torralba. 2020.
\newblock Watch-and-help: A challenge for social perception and human-ai
  collaboration.
\newblock In \emph{International Conference on Learning Representations}.

\bibitem[{Raman et~al.(2022)Raman, Cohen, Rosen, Idrees, Paulius, and
  Tellex}]{raman:neurips22fmdm}
Shreyas~Sundara Raman, Vanya Cohen, Eric Rosen, Ifrah Idrees, David Paulius,
  and Stefanie Tellex. 2022.
\newblock \href
  {http://www.cs.utexas.edu/users/ai-labpub-view.php?PubID=127989} {Planning
  with large language models via corrective re-prompting}.

\bibitem[{Savva et~al.(2019)Savva, Kadian, Maksymets, Zhao, Wijmans, Jain,
  Straub, Liu, Koltun, Malik et~al.}]{savva2019habitat}
Manolis Savva, Abhishek Kadian, Oleksandr Maksymets, Yili Zhao, Erik Wijmans,
  Bhavana Jain, Julian Straub, Jia Liu, Vladlen Koltun, Jitendra Malik, et~al.
  2019.
\newblock Habitat: A platform for embodied ai research.
\newblock In \emph{Proceedings of the IEEE/CVF international conference on
  computer vision}, pages 9339--9347.

\bibitem[{Sellam et~al.(2020)Sellam, Das, and Parikh}]{sellam2020bleurt}
Thibault Sellam, Dipanjan Das, and Ankur Parikh. 2020.
\newblock Bleurt: Learning robust metrics for text generation.
\newblock In \emph{Proceedings of the 58th Annual Meeting of the Association
  for Computational Linguistics}, pages 7881--7892.

\bibitem[{Shah et~al.(2022)Shah, Osinski, Ichter, and Levine}]{shah2022lmnav}
Dhruv Shah, Blazej Osinski, Brian Ichter, and Sergey Levine. 2022.
\newblock \href {http://arxiv.org/abs/2207.04429} {Lm-nav: Robotic navigation
  with large pre-trained models of language, vision, and action}.

\bibitem[{Shen et~al.(2021)Shen, Xia, Li, Mart{\'\i}n-Mart{\'\i}n, Fan, Wang,
  P{\'e}rez-D’Arpino, Buch, Srivastava, Tchapmi et~al.}]{shen2021igibson}
Bokui Shen, Fei Xia, Chengshu Li, Roberto Mart{\'\i}n-Mart{\'\i}n, Linxi Fan,
  Guanzhi Wang, Claudia P{\'e}rez-D’Arpino, Shyamal Buch, Sanjana Srivastava,
  Lyne Tchapmi, et~al. 2021.
\newblock igibson 1.0: A simulation environment for interactive tasks in large
  realistic scenes.
\newblock In \emph{2021 IEEE/RSJ International Conference on Intelligent Robots
  and Systems (IROS)}, pages 7520--7527. IEEE.

\bibitem[{Shridhar et~al.(2019)Shridhar, Thomason, Gordon, Bisk, Han, Mottaghi,
  Zettlemoyer, and Fox}]{2019ALFRED}
Mohit Shridhar, Jesse Thomason, Daniel Gordon, Yonatan Bisk, Winson Han,
  Roozbeh Mottaghi, Luke Zettlemoyer, and Dieter Fox. 2019.
\newblock Alfred: A benchmark for interpreting grounded instructions for
  everyday tasks.

\bibitem[{Song et~al.(2023)Song, Wu, Washington, Sadler, Chao, and
  Su}]{song2023llmplanner}
Chan~Hee Song, Jiaman Wu, Clayton Washington, Brian~M. Sadler, Wei-Lun Chao,
  and Yu~Su. 2023.
\newblock Llm-planner: Few-shot grounded planning for embodied agents with
  large language models.
\newblock In \emph{Proceedings of the IEEE/CVF International Conference on
  Computer Vision (ICCV)}.

\bibitem[{Srivastava et~al.(2021)Srivastava, Li, Lingelbach, Martín-Martín,
  Xia, Vainio, Lian, Gokmen, Buch, and Liu}]{2021BEHAVIOR}
Sanjana Srivastava, Chengshu Li, Michael Lingelbach, Roberto Martín-Martín,
  Fei Xia, Kent Vainio, Zheng Lian, Cem Gokmen, Shyamal Buch, and C.~Karen Liu.
  2021.
\newblock Behavior: Benchmark for everyday household activities in virtual,
  interactive, and ecological environments.

\bibitem[{Team et~al.(2023)Team, Anil, Borgeaud, Wu, Alayrac, Yu, Soricut,
  Schalkwyk, Dai, Hauth et~al.}]{team2023gemini}
Gemini Team, Rohan Anil, Sebastian Borgeaud, Yonghui Wu, Jean-Baptiste Alayrac,
  Jiahui Yu, Radu Soricut, Johan Schalkwyk, Andrew~M Dai, Anja Hauth, et~al.
  2023.
\newblock Gemini: a family of highly capable multimodal models.
\newblock \emph{arXiv preprint arXiv:2312.11805}.

\bibitem[{Wang et~al.(2023)Wang, Xie, Jiang, Mandlekar, Xiao, Zhu, Fan, and
  Anandkumar}]{wang2023voyager}
Guanzhi Wang, Yuqi Xie, Yunfan Jiang, Ajay Mandlekar, Chaowei Xiao, Yuke Zhu,
  Linxi Fan, and Anima Anandkumar. 2023.
\newblock Voyager: An open-ended embodied agent with large language models.
\newblock \emph{arXiv preprint arXiv:2305.16291}.

\bibitem[{Xi et~al.(2023)Xi, Chen, Guo, He, Ding, Hong, Zhang, Wang, Jin, Zhou
  et~al.}]{xi2023rise}
Zhiheng Xi, Wenxiang Chen, Xin Guo, Wei He, Yiwen Ding, Boyang Hong, Ming
  Zhang, Junzhe Wang, Senjie Jin, Enyu Zhou, et~al. 2023.
\newblock The rise and potential of large language model based agents: A
  survey.
\newblock \emph{arXiv preprint arXiv:2309.07864}.

\bibitem[{Xia et~al.(2018)Xia, Zamir, He, Sax, Malik, and
  Savarese}]{xia2018gibson}
Fei Xia, Amir~R Zamir, Zhiyang He, Alexander Sax, Jitendra Malik, and Silvio
  Savarese. 2018.
\newblock Gibson env: Real-world perception for embodied agents.
\newblock In \emph{Proceedings of the IEEE conference on computer vision and
  pattern recognition}, pages 9068--9079.

\bibitem[{Zhang and Kordjamshidi(2023)}]{zhang2023vln}
Yue Zhang and Parisa Kordjamshidi. 2023.
\newblock Vln-trans, translator for the vision and language navigation agent.
\newblock In \emph{The 61st Annual Meeting of the Association for Computational
  Linguistics (ACL-2023)}.

\end{thebibliography}
\bibliographystyle{acl_natbib}

\appendix

\clearpage
\newpage

\begin{figure}[!ht]
\centering
\includegraphics[scale=0.53, trim={0.5cm 0 0.5cm 0}]{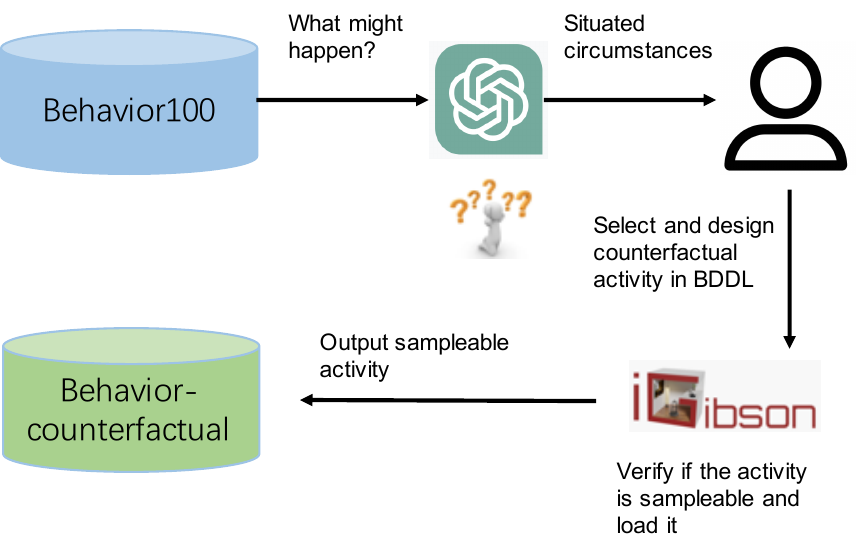}
\caption{Example of prompting LLM for counterfactual activity circumstances.}
%\vskip-1.5em
\label{fig:append 2}
\end{figure}

\section{Annotation Pipeline of Counterfactual Activities}
\label{Appendix:1}

\begin{figure*}[!ht]
\centering
\includegraphics[scale=0.55, trim={0.5cm 0 0.5cm 0}]{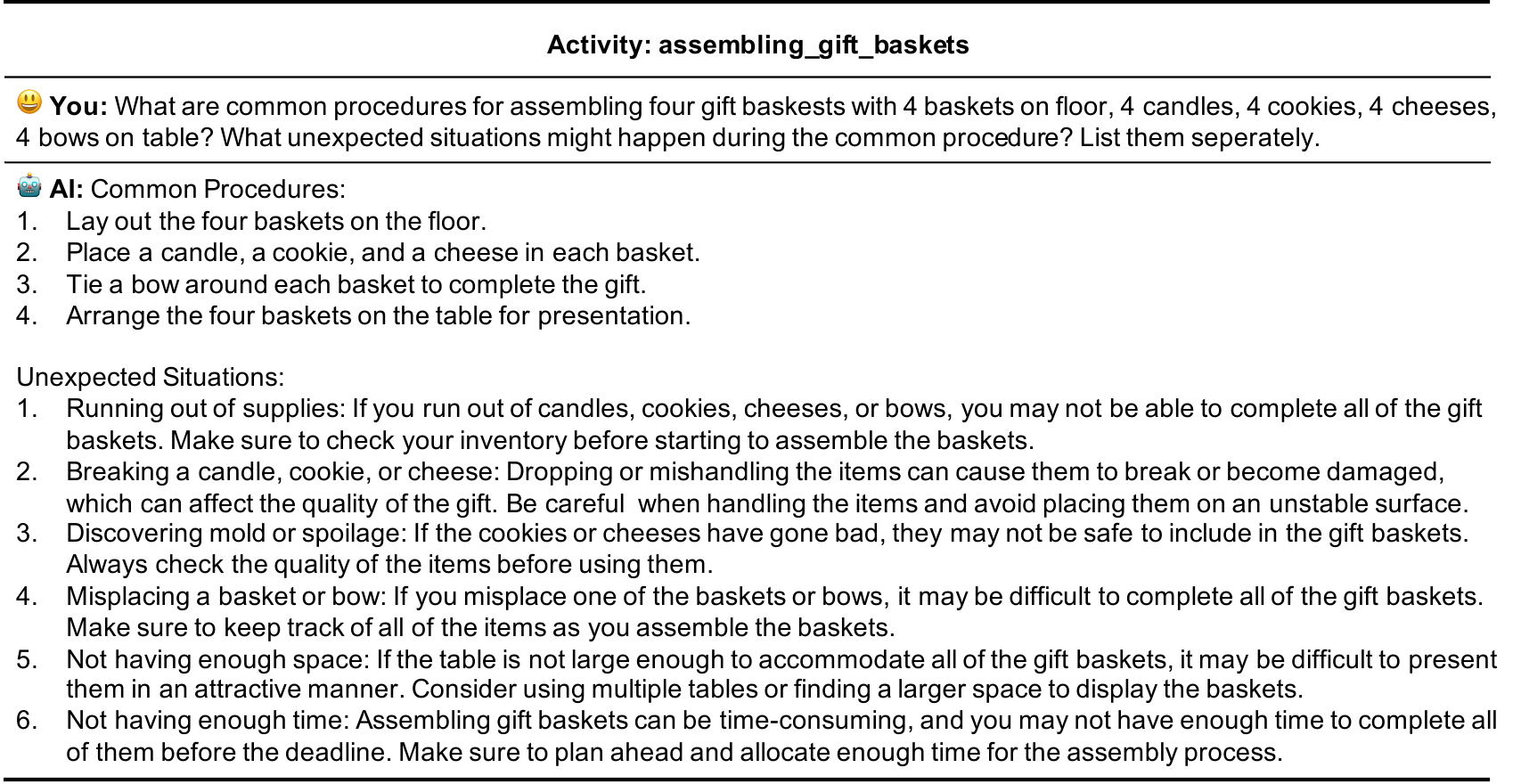}
\caption{Example of prompting LLM for counterfactual activities.}
%\vskip-1.5em
\label{fig:append 1}
\end{figure*}

The prompting example of acquiring situated circumstances is presented in Figure \ref{fig:append 1}. After getting the result, human check on the results and select circumstances that are suitable for defining in the BDDL form and loaded in the simulator. In this example, for circumstance 3, cookies or cheese might gone bad. The simulator can not model states of cookie as ``going bad'' but state of ``burnt''. Therefore, the counterfactual activity is defined with burnt cookies. A new BDDL counterfactual activity definition is constructed. The following natural task description from ChatGPT and image collecting process from iGibson2 are same as normal activities.

As for the counterfactual plan, since burnt cookies are also not safe or suitable to be as gifts, two of the baskets will not be assembled with cookies finally. Therefore, the object state results in a plan change. The full pipeline is presented in Figure \ref{fig:append 2}.

\section{Examples of Counterfactual Activities}
\label{Appendix:2}
\begin{figure}[!ht]
\centering
\includegraphics[scale=0.6, trim={0.5cm 0 0.5cm 0}]{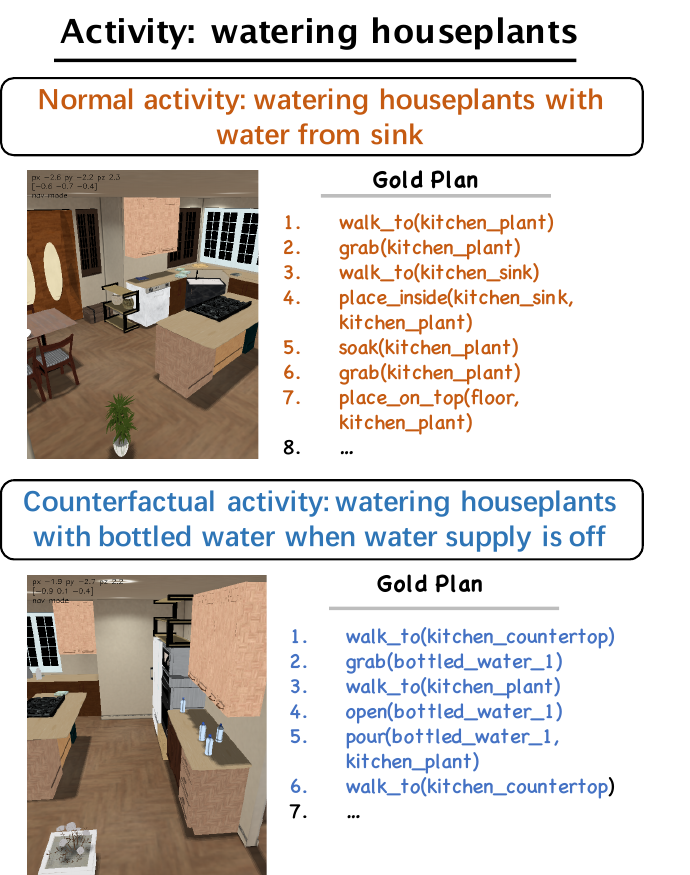}
\caption{Example of normal activity and counterfactual activity.}
%\vskip-1.5em
\label{fig:counterfactual2}
\end{figure}

\begin{figure}[!ht]
\centering
\includegraphics[scale=0.6, trim={0.5cm 0 0.5cm 0}]{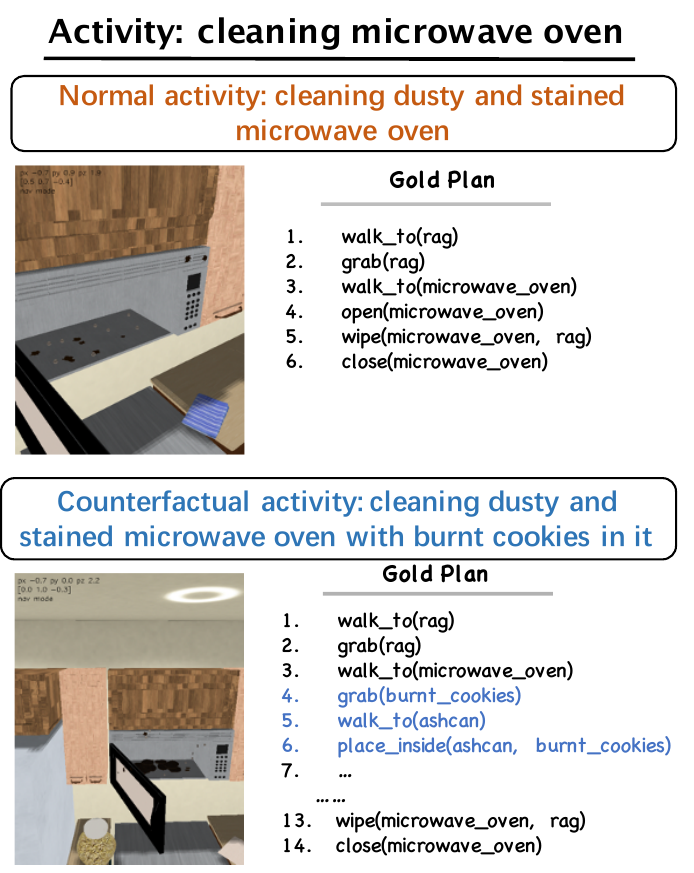}
\caption{Example of normal activity and situated activity.}
%\vskip-1.5em
\label{fig:counterfactual3}
\end{figure}

Examples of counterfactual activities designing on object functions and event causality are shown in Figure \ref{fig:counterfactual2} and \ref{fig:counterfactual3}. 

In Figure \ref{fig:counterfactual2}, when water supply is off, the watering source is bottled water on the countertop. In this activity, the function of bottled water is considered to replace original water source from kitchen sink faucet.  

In Figure \ref{fig:counterfactual3}, before cleaning the microwave oven as common procedures, the agent should first clear the burnt cookies in the microwave oven. In this activity, the clearing burnt cookie event is supposed to  happen before cleaning the microwave oven.

\begin{table*}[]
\begin{tabular}{ccccccc}
\toprule
\multirow{2}{*}{Gemini-pro} & \multicolumn{3}{c}{Normal activity} & \multicolumn{3}{c}{counterfactual activity} \\
     &  Correctness &  Commonsense & Avg Len & Correctness &  Commonsense & Avg Len \\
\midrule
 w image    &   45.0 & 50.0 & 23.75  & 20.0 &  30.0  & 33.30  \\
 w/o image & 32.5 & 40.0 & 48.48  &  15.0 & 25.0 &   60.95 \\
\bottomrule   
\end{tabular}
\caption{Details of ablation study of normal and counterfactual activities with Gemini-Pro-1.5. }
\label{tab:abla-detail}
\end{table*}

\section{VLMs}
\label{Appendix:3}
Gemini-Pro is from a family of hightly capable multimodal models developed at Google \cite{team2023gemini}, which also contains Gemini-Ultra and Gemini-Nano.

Claude 3 is a new family of large multimodal models from Anthropic AI, including Haiku, Sonnet, and Opus\footnote{https://www.anthropic.com/news/claude-3-family}.

GPT-4V \cite{achiam2023gpt} is a large multimodal model capable of processing image and text inputs and producing text outputs, built by OpenAI.

\section{Example of Errors}
\label{Appendix:4}
For each of the six error types, one example sampled from the VLM prompting results is presented in Table \ref{appendx:error}. 

\begin{table*}[!ht]
\resizebox{\textwidth}{!}{
\begin{tabular}{lcccc}
\toprule
Error Type   & Activity   & Image Input & VLM Plan   & Gold Plan \\
\midrule
\begin{tabular}[c]{@{}l@{}}missing\\ actions\end{tabular}                      & \begin{tabular}[c]{@{}c@{}}cleaning \\ freezer\end{tabular}  &    
    \begin{minipage}{0.2\textwidth} \includegraphics[width=25mm]{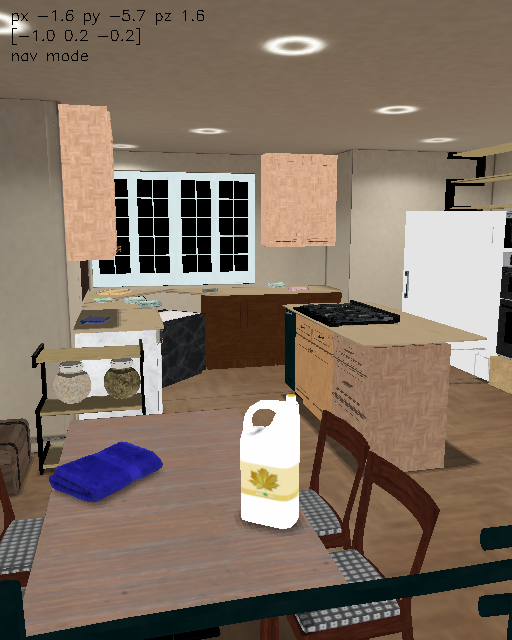} 
 \end{minipage} & 
 
  \begin{tabular}[c]{@{}c@{}}...\\ walk\_to(refrigerator)\\ open(refrigerator)\\ grab(food\_item1)\\ place\_inside(sink, food\_item1)\\ ...\end{tabular}          & \begin{tabular}[c]{@{}c@{}}...\\ walk\_to(refrigerator)\\ open(refrigerator)\\ grab(food\_item1)\\ walk\_to(sink)\\ place\_inside(sink, food\_item1)\\ ...\end{tabular}       \\
\midrule
\begin{tabular}[c]{@{}l@{}}mistake of \\ object \\ property/\\ function\end{tabular} & \begin{tabular}[c]{@{}c@{}}packing food \\ for work\end{tabular}                                                               &  
\begin{minipage}{0.2\textwidth} \includegraphics[width=25mm]{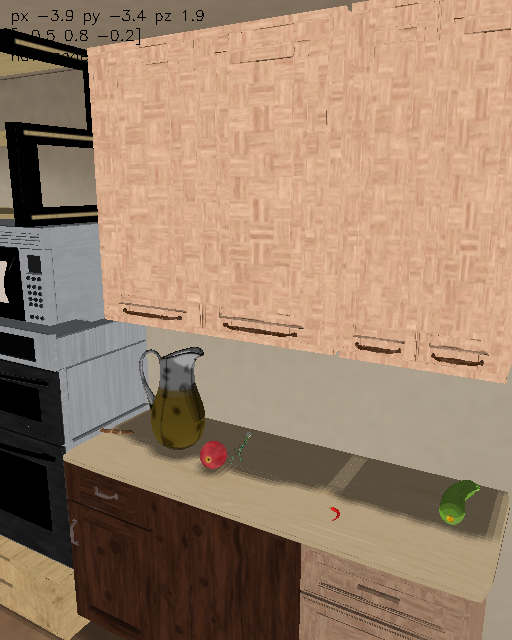} 
 \end{minipage} 
& \begin{tabular}[c]{@{}c@{}}...\\ walk\_to(countertop)\\ grab(chives)\\ place\_inside(carton, chives)\\ grab(cinnamon)\\ place\_inside(carton, cinnamon)\\ ...\end{tabular}                                                                                                 & \begin{tabular}[c]{@{}c@{}}...\\ walk\_to(countertop)\\ grab(chives)\\ place\_inside(carton, chives)\\ grab(chili)\\ place\_inside(carton, chili)\\ ...\end{tabular}                                                                                                          \\
\midrule
\begin{tabular}[c]{@{}l@{}}mistake of \\ event \\ cause/result\end{tabular}   & \begin{tabular}[c]{@{}c@{}}cleaning \\ sneakers\end{tabular}                                                            &     
\begin{minipage}{0.2\textwidth} \includegraphics[width=25mm]{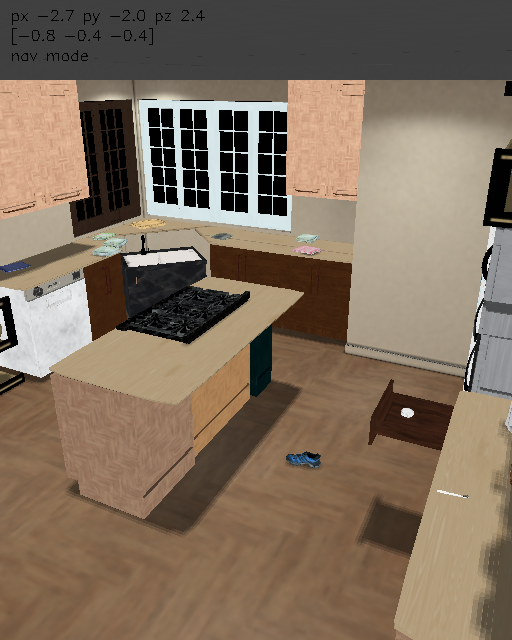} 
 \end{minipage} 
 & 
 \begin{tabular}[c]{@{}c@{}}...\\ walk\_to(sink)\\ switch\_on(faucet)\\ place\_inside(sink, soap)\\ switch\_off(faucet)\\ walk\_to(living\_room)\\ walk\_to(sneaker1)\\ clean(sneaker1, soap, brush)\\ ...\end{tabular}                                                     & \begin{tabular}[c]{@{}c@{}}...\\ walk\_to(sink)\\ switch\_on(faucet)\\ place\_inside(sink, soap)\\ switch\_off(faucet)\\ walk\_to(living\_room)\\ walk\_to(sneaker1)\\ grab(sneaker\_1)\\ walk\_to(sink)\\ place\_inside(sink, sneaker1)\\ clean(sneaker1)\\ ...\end{tabular} \\
\midrule
\begin{tabular}[c]{@{}l@{}}incorrect \\ form\end{tabular}            & \begin{tabular}[c]{@{}c@{}}brush lint \\ off clothing\end{tabular}                                                       &       \begin{minipage}{0.2\textwidth} \includegraphics[width=25mm]{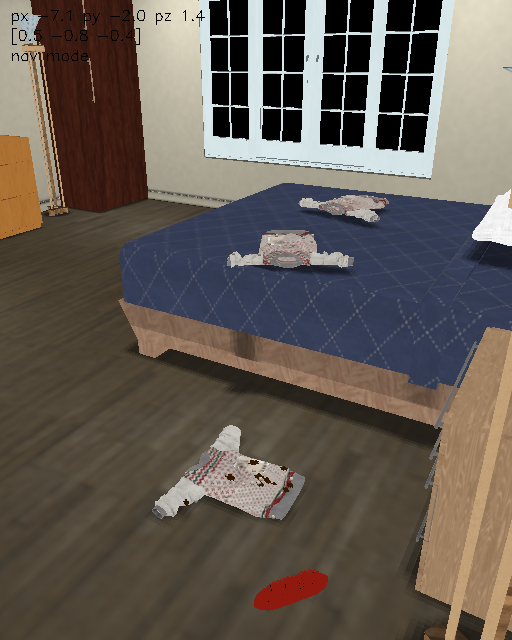} 
 \end{minipage} 
 & \begin{tabular}[c]{@{}c@{}}...\\ walk\_to(scrub\_brush)\\ grab(scrub\_brush)\\ grab(sweater1)\\ (brush sweater1)\\ ...\end{tabular}                 & \begin{tabular}[c]{@{}c@{}}...\\ grab(scrub\_brush)\\ grab(sweater1\_on\_floor)\\ brush(sweater1\_on\_floor, scrub\_brush)\\ ...\end{tabular}                                                                      \\
\midrule
\begin{tabular}[c]{@{}l@{}}hallucination \\ of extra tools\end{tabular}         & \begin{tabular}[c]{@{}c@{}}watering \\ houseplants\end{tabular}                                                           &      
\begin{minipage}{0.2\textwidth} \includegraphics[width=25mm]{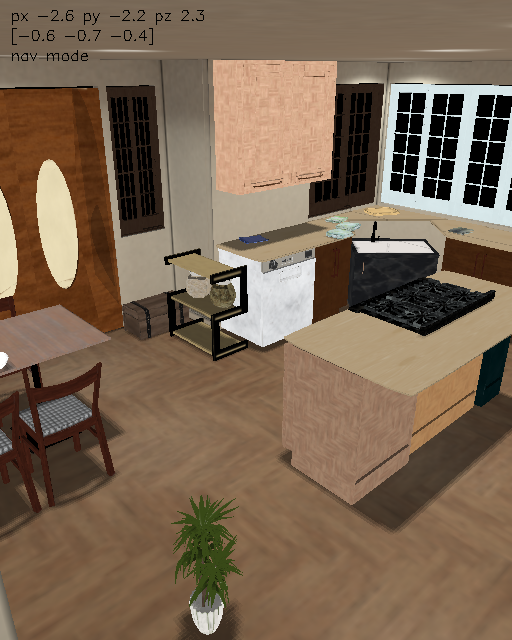} 
 \end{minipage}
 & \begin{tabular}[c]{@{}c@{}}...\\ walk\_to(dining\_room\_plant\_1)\\ grab(watering\_can)\\ pour(watering\_can, dining\_room\_plant\_1)\\ place(watering\_can)\\ walk\_to(dining\_room\_plant\_2)\\ ...\end{tabular}                                                         & \begin{tabular}[c]{@{}c@{}}...\\ walk\_to(kitchen)\\ walk\_to(sink)\\ place\_inside(sink, dining\_room\_plant1)\\ place\_inside(sink, dining\_room\_plant2)\\ soak(dining\_room\_plant1)\\ soak(dining\_room\_plant2)\\ ...\end{tabular}                                      \\
\midrule
\begin{tabular}[c]{@{}l@{}}incorrect image \\ understanding\end{tabular}       & \begin{tabular}[c]{@{}c@{}}putting dishes \\ away after \\ cleaning\end{tabular}   &   

\begin{minipage}{0.2\textwidth} \includegraphics[width=25mm]{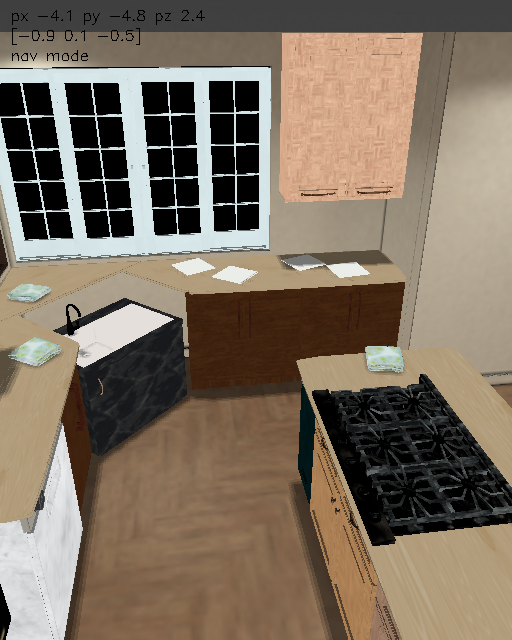} 
 \end{minipage} 
 
& \begin{tabular}[c]{@{}c@{}}walk\_to(countertop\_1)\\ grab(plate\_1)\\ walk\_to(cabinet)\\ open(cabinet)\\ place\_inside(cabinet, plate\_1)\\ walk\_to(countertop\_2)\\ grab(plate\_2)\\ walk\_to(cabinet)\\ place\_inside(cabinet, plate\_2)\\ close(cabinet)\end{tabular} & \begin{tabular}[c]{@{}c@{}}...\\ walk\_to(kitchen)\\ walk\_to(countertop\_1)\\ grab(plate\_1)\\ grab(plate\_2)\\ grab(plate\_3)\\ grab(plate\_4)\\ walk\_to(countertop\_2)\\ grab(plate\_5)\\ grab(plate\_6)\\ grab(plate\_7)\\ grab(plate\_8)\\ ...\end{tabular}   \\
  \bottomrule
\end{tabular}
}
\caption{Error examples.}
\label{appendx:error}
\end{table*}

\section{Ablation Study}
\label{appendix:abla}
As presented in Table \ref{tab:abla-detail}, counterfactual activities have the same trend with normal activity, while the scores of normal activities drop greater than counterfactual activities. 

\section{Evaluation of open-source VLMs on ActPlan-1K}
We have conducted experiments on typical VLMs like VILA \cite{lin2024vila}. However, current open-source VLMs can not generate structured results as required(e.g., walk\_to(table). One example output for activating assembling basket is as follows:

\textit{1. Walk to the table. 2. Place a cookie on top of the bowl. 3. Grab a candle from the floor. 4. Place the candle next to the cookie on the table. 5. Open the cheese container and place a cheese inside. 6. Close the cheese container. 7. Walk to the floor. 8. Place a cookie on top of the bowl. 9. Grab a candle from the floor. … … 38. Walk to the floor. 39. Place a cookie on top of the bowl. 40. Grab a candle from the.}

Therefore, we have not reported the results on open-source VLMs.

\section{Human Annotators}
For the human annotation in gold plans and human scoring of correctness and commonsense satisfaction, we train three PhD students from CSE department with the requirements illustrated in the paper. By checking annotation results from the students, the overall annotation quality is good on the above tasks.

\end{document}